\documentclass{article}

\usepackage[left=1in, right=1in, top=1in, bottom=1in]{geometry}

\usepackage[square,sort,comma,numbers]{natbib}

\usepackage{microtype}
\usepackage{graphicx}
\usepackage{subfigure}
\usepackage{array}
\usepackage{booktabs} 
\usepackage{amsfonts}
\usepackage{amssymb,amsmath,amsthm,bm}

\usepackage{soul}
\usepackage{lmodern}
\usepackage{booktabs}
\usepackage{threeparttable}
\usepackage{mathtools}
\usepackage{tablefootnote}

\usepackage{cuted}
\usepackage{lipsum, color}

\usepackage{verbatim}

\usepackage{hyperref}

\usepackage{listings}
\usepackage{color}

\usepackage{color}

\definecolor{mygreen}{rgb}{0,0.6,0}
\definecolor{mygray}{rgb}{0.5,0.5,0.5}
\definecolor{mymauve}{rgb}{0.58,0,0.82}

\usepackage{algorithm}
\usepackage[noend]{algpseudocode}

\usepackage{soul}

\title{Adversarial Defense via the Data-Dependent Activation, Total Variation Minimization, and Adversarial Training}

\author{
Bao Wang\footnote{Please correspond to: wangbaonj@gmail.com}  \\
  Scientific Computing and Imaging (SCI) Institute\\
  University of Utah, Salt Lake City, UT, USA\\
  \and
  Alex Lin\\
  Department of Mathematics\\ 
  University of California, Los Angeles\\
  \and
  Penghang Yin \\
  Department of Mathematics\\ 
  University of California, Los Angeles\\
  \and
  Wei Zhu\\
  Department of Mathematics\\ 
  Duke University, Durham\\
  \and
  Andrea L. Bertozzi \\
  Department of Mathematics\\
  University of California, Los Angeles\\
  \and
  Stanley J. Osher \\
  Department of Mathematics\\
  University of California, Los Angeles\\
}

\begin{document}

\maketitle

\begin{abstract}
We improve the robustness of Deep Neural Net (DNN) to adversarial attacks by using an interpolating function as the output activation. This data-dependent activation remarkably improves both the generalization and robustness of DNN. In the CIFAR10 benchmark, we raise the robust accuracy of the adversarially trained ResNet20 from $\sim 46\%$ to $\sim 69\%$ under the state-of-the-art Iterative Fast Gradient Sign Method (IFGSM) based adversarial attack. When we combine this data-dependent activation with total variation minimization on adversarial images and training data augmentation, we achieve an improvement in robust accuracy by 38.9$\%$ for ResNet56 under the strongest IFGSM attack. Furthermore, We provide an intuitive explanation of our defense by analyzing the geometry of the feature space.
\end{abstract}

\section{Introduction}
The adversarial vulnerability \cite{Szegedy:2013} of Deep Neural Nets (DNNs) threaten their applicability in security critical tasks, e.g., autonomous cars \cite{Akhtar:2018}, robotics \cite{Giusti:2016Drones}, DNN-based malware detection systems \cite{PapernotSecurity:2016,PapernotMalware:2016}. Since the pioneering work by Szegedy et al. \cite{Szegedy:2013}, many advanced adversarial attacks have been devised to generate imperceptible perturbations to fool the DNN \cite{Goodfellow:2014AdversarialTraining,PapernotAttack:2016,CWAttack:2016,Wu:2018,Ilyas:2018,Athalye:2018B,Dong_2018_CVPR}. Not only are adversarial attacks successful in white-box attacks, i.e., when the adversary has access to the DNN parameters, but they are also successful in black-box attacks, i.e., without access to network parameters. Adversarial attacks are transferable in the sense that a perturbed image meant to be misclassified by one DNN also has a significant chance to be misclassified by another DNN \cite{DBLP:journals/corr/PapernotMG16}. Due to this transferability, adversaries can attack DNN without knowing the network parameters (i.e. blackbox) \cite{LiuYanpei:2016,Brendel:2017}. There even exist universal perturbations that can imperceptibly perturb any image and cause misclassification for any given network \cite{Moosavi-Dezfooli_2017_CVPR}. And recently, there has been much work on defending against these universal perturbations \cite{Akhtar_2018_CVPR}.

In this work, we defend against adversarial attacks by replacing the commonly used output activation of DNN with a manifold-interpolating function. Together with the Projected Gradient Descent (PGD) adversarial training \cite{Madry:2018}, Total Variation Minimization (TVM), and training data augmentation, we show state-of-the-art results for adversarial defense on the CIFAR10 benchmark.

\subsection{Related Work}
Defensive distillation was recently proposed to increase the robustness of DNN \cite{PapernotDistillationDefense:2016}, and a related approach \cite{tramer2018ensemble} cleverly modifies the training data to increase robustness against black-box attacks and adversarial attacks in general. To counter adversarial perturbations, Guo et al. \cite{ChuanGuo:2018}, proposed to use image transformation, e.g., bit-depth reduction, JPEG compression, TVM, and image quilting. A similar idea of denoising the input was later explored in \cite{DBLP:journals/corr/abs-1802-06806}, where the authors divide the input into patches, denoise each patch, and then reconstruct the image. These input transformations are intended to be non-differentiable, thus making adversarial attacks more difficult, especially for gradient-based attacks. Another denoising approach is introduced by Liao et al. \cite{Liao_2018_CVPR}, where they proposed a High-level Representation Guided (HGD) denoiser -- the idea is that while perturbations seem small in the original and adversarial images, these perturbations are amplified in higher representations. Transformation-based defenses have also been proposed by Xie et al. \cite{ie:2018}, and Luo et al. \cite{DBLP:journals/corr/LuoBRPZ15}. Song et al. \cite{Song:2018}, noticed that small adversarial perturbations shift the distribution of adversarial images far from the distribution of clean images. Therefore, they proposed to purify the adversarial images by PixelDefend. And Prakash et al. \cite{Prakash_2018_CVPR}, also seek to examine image statistics in order to construct an adversarial defense -- in their work, they introduce Pixel Deflection where they force images to match statistics of natural images. Lee et al. \cite{2018arXiv180703888L}, have also used the distribution of images to detect adversarial examples. Adversarial training is another family of defense methods to enhance the stability of DNN \cite{Goodfellow:2014AdversarialTraining,Madry:2018,Na:2018}. In particular, the PGD adversarially trained DNN achieves state-of-the-art resistance to the available attacks \cite{Madry:2018}. GANs are also employed for adversarial defense \cite{Samangouei:2018}. In \cite{Athalye:2018}, the authors proposed an approximated gradient to attack the defenses that are based on the obfuscated gradient.

Instead of using the softmax function as DNN's output activation, Wang et al. \cite{BaoWang:2018NIPS,BaoWang:2019}, utilized a class of non-parametric interpolating functions. This is a combination of both deep and manifold learning which causes the DNN to utilize the geometric information of the training data sufficiently. The authors show a significant amount of generalization accuracy improvement, and the results are more stable when one only has a limited amount of training data. Recently, Wang et al. \cite{wang2019resnets} modeled ResNet as a transport equation, and they proposed an Feynman-Kac formalism principled adversarial robust DNN.

\subsection{Organization}
We organize this paper as follows: In section~\ref{section:DNN:Interpolation}, we overview the DNN with a graph Laplacian-based high dimensional interpolating activation function. In section~\ref{section:Attacks}, we present a few adversarial attacks that will be used as benchmarks for this work. In section~\ref{section:Interpolate:TVM}, we elaborate on adversarial defense via interpolating activation together with TVM. In section~\ref{section:PGD}, we further study the robustness of PGD adversarially trained DNN with interpolating activation. This paper ends up with concluding remarks.

\section{DNN with Data-Dependent Activation}\label{section:DNN:Interpolation}
\begin{figure}[!ht]
\centering
\begin{tabular}{cc}
\includegraphics[width=0.42\columnwidth]{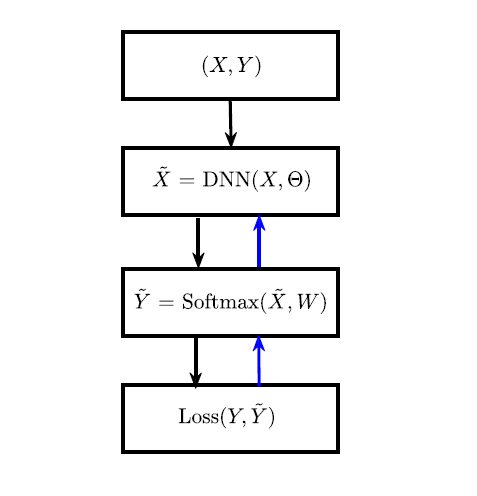}&
\includegraphics[width=0.42\columnwidth]{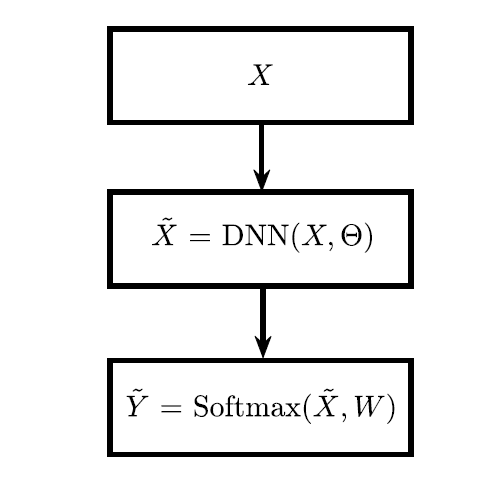}\\
(a)&(b)\\
\includegraphics[width=0.51\columnwidth]{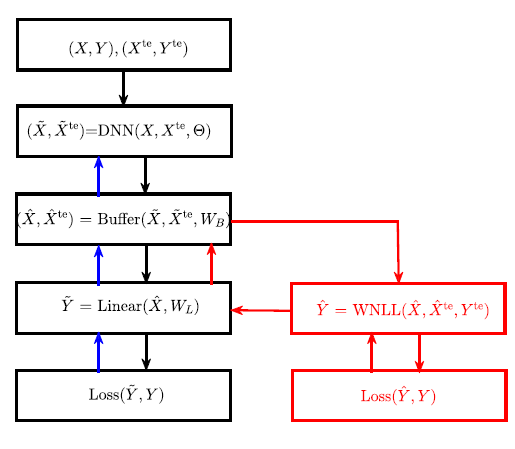}&
\includegraphics[width=0.41\columnwidth]{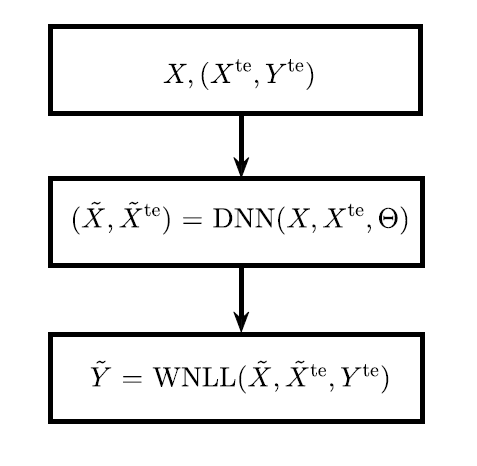}\\
(c)&(d)\\
\end{tabular}
\caption{Training and testing procedures of the DNN with softmax and WNLL functions as the output activation layer. (a) and (b) show the training and testing steps for the standard DNN, respectively; (c) and (d) illustrate the training and testing procedure of the WNLL activated DNN, respectively.}
\label{fig:WNLL-DNN-Structure}
\end{figure}

In this section, we summarize the architecture, training, and testing procedures of the DNN with the data-dependent activation \cite{BaoWang:2018NIPS}. For the standard DNN with softmax activation, the training and testing are shown in Fig.~\ref{fig:WNLL-DNN-Structure} (a) and (b), respectively. In the $k$th iteration of training, given a mini-batch of training data $\mathbf{X}, \mathbf{Y}$, we perform:

{\it Forward propagation:} Transform $\mathbf{X}$ into features by the DNN block (a combination of convolutional layers, nonlinearities, etc.), and then feed the output into the softmax activation to obtain the predictions $\tilde{\mathbf{Y}}$, i.e., 
{
$$
\tilde{\mathbf{Y}} = {\rm Softmax}({\rm DNN}(\mathbf{X}, \Theta^{k-1}), \mathbf{W}^{k-1}).
$$
}
Then the loss is computed (e.g., cross entropy) between $\mathbf{Y}$ and $\tilde{\mathbf{Y}}$:
$\mathcal{L} \doteq  \mathcal{L}^{\rm Linear} = {\rm Loss}(\mathbf{Y}, \tilde{\mathbf{Y}})$.

{\it Backpropagation:} Update weights ($\Theta^{k-1}$, $\mathbf{W}^{k-1}$) by gradient descent with learning rate $\gamma$
{
$$\mathbf{W}^{k} = \mathbf{W}^{k-1} - \gamma \frac{\partial \mathcal{L}}{\partial \tilde{\mathbf{Y}}}\cdot \frac{\partial \tilde{\mathbf{Y}}}{\partial \mathbf{W}},
$$

$$\Theta^{k} = \Theta^{k-1} - \gamma \frac{\partial \mathcal{L}}{\partial \tilde{\mathbf{Y}}}\cdot \frac{\partial \tilde{\mathbf{Y}}}{\partial \tilde{\mathbf{X}}}\cdot \frac{\partial \tilde{\mathbf{X}}}{\partial \Theta}.$$
}

Once the model is optimized, with optimal parameters being $(\Theta, \mathbf{W})$,
the predicted labels for testing data $\mathbf{X}$ are
$$
\tilde{\mathbf{Y}} = {\rm Softmax}({\rm DNN}(\mathbf{X}, \Theta), \mathbf{W}).
$$

Wang et al \cite{BaoWang:2018NIPS} proposed to replace the data-agnostic softmax by an interpolating function defined below.

\subsection{Manifold Interpolation - A Harmonic Extension Approach}
Let $\mathbf{X} = \{\mathbf{x}_1, \mathbf{x}_2, \cdots, \mathbf{x}_n\}$ be a set of points on a high dimensional manifold $\mathcal{M}\subset\mathbb{R}^d$ and $\mathbf{X}^{\rm te} = \{\mathbf{x}^{\rm te}_1, \mathbf{x}^{\rm te}_2, \cdots, \mathbf{x}^{\rm te}_m\}$ (``te" for template) be a subset of $\mathbf{X}$ which are labeled with label function $g(\mathbf{x})$ \footnote{
The minimum requirement is that the template data needs to cover all classes. In \cite{BaoWang:2018NIPS}, we show that for an image classification task with $m$ number of different classes, the size of the template needs to be at least $m\log m$. In practice, the size of the template set will not affect the performance much as long as the template set size is more than 1K for CIFAR10 and CIFAR100.}.  We want to interpolate a function $u$ that is defined on $\mathcal{M}$ and can be used to label the entire dataset $\mathbf{X}$. The harmonic extension is a natural approach to find such an interpolating function, which is defined by minimizing the following Dirichlet energy functional
{
\begin{equation}
\label{DirichletEnergy}
\mathcal{E}(u)=\frac{1}{2}\sum_{\mathbf{x}, \mathbf{y}\in \mathbf{X}} w(\mathbf{x}, \mathbf{y})\left(u(\mathbf{x})-u(\mathbf{y})\right)^2,
\end{equation}
}
with the boundary condition
{
$$
u(\mathbf{x})=g(\mathbf{x}), \ \mathbf{x}\in \mathbf{X}^{\rm te},
$$
}
where $w(\mathbf{x}, \mathbf{y})$ is a weight function, chosen to be Gaussian: $w(\mathbf{x}, \mathbf{y})=\exp(-\frac{||\mathbf{x}-\mathbf{y}||^2}{\sigma^2})$ with $\sigma$ being a scaling parameter. The Euler-Lagrange equation for Eq.~(\ref{DirichletEnergy}) is
{
\begin{equation}
\label{EL-Equation}
\begin{cases}
\sum_{\mathbf{y}\in \mathbf{X}} \left(w(\mathbf{x}, \mathbf{y})+w(\mathbf{y}, \mathbf{x})\right)\left(u(\mathbf{x})-u(\mathbf{y})\right)=0 & \hskip -0.2cm \mathbf{x}\in \mathbf{X}/\mathbf{X}^{\rm te}\\
u(\mathbf{x})=g(\mathbf{x}) & \hskip -0.3cm \mathbf{x}\in \mathbf{X}^{\rm te}.
\end{cases}
\end{equation}
}

By solving the linear system Eq.~(\ref{EL-Equation}), we obtain labels $u(\mathbf{x})$ for the unlabeled data $\mathbf{x}\in \mathbf{X}/\mathbf{X}^{\rm te}$. This interpolation becomes invalid when the labeled data is tiny, i.e., $|\mathbf{X}^{\rm te}|\ll |\mathbf{X}
/\mathbf{X}^{\rm te}|$. To resolve this issue, the weights of the labeled data is increased in the Euler-Lagrange equation, which gives

{
\begin{equation}
\label{WNLL}
\begin{cases}
\sum_{\mathbf{y}\in \mathbf{X}} \left(w(\mathbf{x}, \mathbf{y})+w(\mathbf{y}, \mathbf{x})\right)\left(u(\mathbf{x})-u(\mathbf{y})\right)+\\
\left(\frac{|\mathbf{X}|}{|\mathbf{X}^{\rm te}|}-1\right)\sum_{\mathbf{y}\in \mathbf{X}^{\rm te}}w(\mathbf{y}, \mathbf{x})\left(u(\mathbf{x})-u(\mathbf{y})\right)=0 & \hskip -0.2cm \mathbf{x}\in \mathbf{X}/\mathbf{X}^{\rm te}\\
u(\mathbf{x})=g(\mathbf{x}) &\hskip -0.2cm \mathbf{x}\in \mathbf{X}^{\rm te}.
\end{cases}
\end{equation}
}

The solution to Eq.~(\ref{WNLL}) is named weighted nonlocal Laplacian (WNLL), denoted as ${\rm WNLL}(\mathbf{X}, \mathbf{X}^{\rm te}, \mathbf{Y}^{\rm te})$. Shi et al. \cite{WNLL:2018}, showed that WNLL converges to the solution of the high dimensional Laplace-Beltrami equation. For classification, $g(\mathbf{x})$ is the one-hot label for the example $\mathbf{x}$.

\subsection{Training and Testing of the DNN with Data-Dependent Activation Function}
For a standard DNN, we denote the WNLL activated one as DNN-WNLL, e.g., the WNLL activated ResNet20 is denoted as ResNet20-WNLL. In both training and testing of the DNN-WNLL, we need to reserve a small portion of data/label pairs, denoted as $(\mathbf{X}^{\rm te}, \mathbf{Y}^{\rm te})$, to interpolate the label for new data. We name the reserved data $(\mathbf{X}^{\rm te}, \mathbf{Y}^{\rm te})$ as the template. Directly replacing softmax by WNLL has difficulties in back propagation, namely the true gradients $\frac{\partial \mathcal{L}}{\partial \Theta}$ and $\frac{\partial \mathcal{L}}{\partial \mathbf{W}_B}$ (here $\mathcal{L}\doteq \mathcal{L}^{WNLL} = {\rm Loss}(\hat{Y}, Y)$, as shown in Fig.~\ref{fig:WNLL-DNN-Structure}(c)) are difficult to compute since WNLL defines an implicit function. Instead, to train the DNN-WNLL, a proxy via an auxiliary DNN (Fig.~\ref{fig:WNLL-DNN-Structure}(c)) is employed. On top of the original DNN, we add a buffer block (a fully connected layer followed by a ReLU), and followed by two parallel branches, WNLL and linear (fully connected) layers. The auxiliary DNN can be trained by alternating between training the DNN with linear and WNLL activation functions, respectively. When training DNN with WNLL activation function, the training loss of the WNLL activation is backpropped via a straight-through gradient estimator \cite{Athalye:2018,BengioStraightTE:2013}, e.g., in the $k$th iteration, we use the following approximated gradient descent (Eq.~(\ref{ApproxBP})) to update $\mathbf{W}_B$ only (when backpropagating the training loss $\mathcal{L}^{\rm WNLL}$ we freeze the remaining part except for the buffer block, and the other parameters will be updated in training DNN with linear activation function),
{
\begin{eqnarray}
\label{ApproxBP}
\mathbf{W}_B^{k} =  \mathbf{W}_B^{k-1} - \gamma \frac{\partial \mathcal{L}^{\rm WNLL}}{\partial \hat{\mathbf{Y}}}\cdot \frac{\partial \hat{\mathbf{Y}}}{\partial \hat{\mathbf{X}}}\cdot \frac{\partial \hat{\mathbf{X}}}{\partial \mathbf{W}_B} \\ \nonumber
\approx \mathbf{W}_B^{k-1} - \gamma \frac{\partial \mathcal{L}^{\rm Linear}}{\partial \tilde{\mathbf{Y}}}\cdot \frac{\partial \tilde{\mathbf{Y}}}{\partial \hat{\mathbf{X}}}\cdot \frac{\partial \hat{\mathbf{X}}}{\partial \mathbf{W}_B},
\label{eq:bp-wnll}
\end{eqnarray}
}
where $\frac{\partial \mathcal{L}^{\rm Linear}}{\partial \tilde{\mathbf{Y}}}$ and $\frac{\partial \mathcal{L}^{\rm WNLL}}{\partial \hat{\mathbf{Y}}}$ are the gradients computed through  
two different activation functions. In the approximation of Eq.~(\ref{ApproxBP}), we simply replace the value of $\mathcal{L}^{\rm Linear}$ with that of $\mathcal{L}^{\rm WNLL}$, which allows us to compute the value of $\frac{\partial \mathcal{L}^{\rm WNLL}}{\partial \hat{\mathbf{Y}}}$ by leveraging the computational graph of DNN with linear activation. The detailed training procedure can be found in \cite{BaoWang:2018NIPS}.

At test time, we remove the linear activation from the neural net and use the DNN and buffer blocks together with WNLL to classify new data (Fig. \ref{fig:WNLL-DNN-Structure} (d)). Here for simplicity, we merge the buffer block to the DNN block. For a given set of testing data $\mathbf{X}$, and the labeled template $\{(\mathbf{X}^{\rm te}, \mathbf{Y}^{\rm te})\}$, the predicted labels for $\mathbf{X}$ is given by
{
$$
\tilde{\mathbf{Y}} = {\rm WNLL}({\rm DNN}(\mathbf{X}, \mathbf{X}^{\rm te}, \Theta), \mathbf{Y}^{\rm te}).
$$
}

\subsection{Computational Complexity of DNN with Data-Dependent Activation}
Using WNLL activation will lead to some extra computational overhead, which comes from the nearest neighbor searching and solving a system of linear equations. We following the same training procedure as that used in \cite{BaoWang:2018NIPS} to train ResNet20.
In Table~\ref{Numerical-Compare-Complexity}, we list the training and test time on a single Titan Xp GPU for ResNet20 on CIFAR10.
\begin{table}[!ht]
\renewcommand{\arraystretch}{1.3}
\centering
\caption{Running time and GPU memory for ResNet20 with two different activation functions.}
\label{Numerical-Compare-Complexity}
\centering
\begin{tabular}{cccc}
\hline
 & Training time & Testing time & Memory\\
\hline
ResNet20 & 3925.6 (s)  & 0.657 (s) & 1007 (MB)\\
ResNet20-WNLL    & 7378.4 (s) & 14.09 (s)  & 1563 (MB)\\
\hline
\end{tabular}
\end{table}

\section{Adversarial Attacks}\label{section:Attacks}
We consider three benchmark attacks: the Fast Gradient Sign Method (FGSM) \cite{Goodfellow:2014AdversarialTraining}, Iterative FGSM (IFGSM) \cite{Kurakin:2016}, and Carlini-Wagner's $L_2$ (CW-L2) \cite{CWAttack:2016} attack. We denote the classifier defined by the DNN as $\tilde{y} = f(\theta, \mathbf{x})$ for a given instance ($\mathbf{x}$, $y$). FGSM searchs the adversarial image $\mathbf{x}'$ with a bounded perturbation by maximizing the loss $\mathcal{L}(\mathbf{x}', y) \doteq \mathcal{L}(f(\theta, \mathbf{x}'), y)$, subject to the $l_\infty$ perturbation constraint $||\mathbf{x}'-\mathbf{x}||_\infty \leq \epsilon$ with $\epsilon$ being the attack strength. We can approximately solve this constrained optimization problem by using the first order approximation of the loss function i.e., $\mathcal{L}(\mathbf{x}', y) \approx \mathcal{L}(\mathbf{x}, y) + \nabla_\mathbf{x}\mathcal{L}(\mathbf{x}, y)^T \cdot (\mathbf{x}'-\mathbf{x})$. Under this approximation, the optimal adversarial image is
\begin{equation}
\label{FGSM}
\mathbf{x}'=\mathbf{x} + \epsilon \, {\rm sign} \cdot \left( \nabla_\mathbf{x}\mathcal{L}(\mathbf{x}, y) \right).
\end{equation}

IFGSM iterates FGSM to generate the enhanced attack, i.e.,
\begin{equation}
\label{IFGSM-2}
\mathbf{x}^{(m)} = {\rm Clip}_{\mathbf{x}, \epsilon}\left\{\mathbf{x}^{(m-1)} + \alpha \cdot {\rm sign} \left( \nabla_{\mathbf{x}} \mathcal{L}(\mathbf{x}^{(m-1)}, y) \right)\right\},
\end{equation}
where $m=1, \cdots, M$, $\mathbf{x}^{(0)}=\mathbf{x}$ and $\mathbf{x}'=\mathbf{x}^{(M)}$, with $M$ be the number of iterations. $\alpha$ is the step size used in each iteration, and ${\rm Clip}_{\mathbf{x}, \epsilon}$ clips the update to be within an $\epsilon$-ball centered at $\mathbf{x}$ in $l_\infty$-norm.

Moreover, we consider the attack due to Carlini and Wagner. For a given image-label pair $(\mathbf{x}, y)$, and $\forall t\neq y$, CW-L2 searches the adversarial image that will be classified to class $t$ by solving the optimization problem

\begin{equation}
\label{cwl2-eq1}
\min_{\delta} ||\delta||_2^2,
\end{equation}
subject to
$$
f(\mathbf{x}+\delta) = t, \; \mathbf{x}+\delta \in [0, 1]^n,
$$
where $\delta$ is the adversarial perturbation (for simplicity, we ignore the dependence of $\theta$ in $f$). 

The equality constraint in Eq.~(\ref{cwl2-eq1}) is hard to handle, so Carlini et al. considered the surrogate
\begin{equation}
\label{cwl2-eq2}
g(\mathbf{x}) = \max\left(\max_{i\neq t}(Z(\mathbf{x})_i) - Z(\mathbf{x})_t , 0\right),
\end{equation}

where $Z(\mathbf{x})$ is the logit vector for an input $\mathbf{x}$, i.e., output of DNN before the output layer. $Z(\mathbf{x})_i$ is the logit value corresponding to class $i$. It is easy to see that $f(\mathbf{x}+\delta)=t$ is equivalent to $g(\mathbf{x}+\delta)\leq 0$. Therefore, the problem in Eq.~(\ref{cwl2-eq1}) can be reformulated as

\begin{equation}
\label{cwl2-eq4}
\min_{\delta} ||\delta||_2^2 + c \cdot g(\mathbf{x}+\delta),
\end{equation}
subject to
$$
\mathbf{x}+\delta \in [0, 1]^n,
$$
where $c\geq 0$ is the Lagrangian multiplier. 

By letting $\delta = \frac{1}{2}\left(\tanh(\mathbf{w})+1\right)-\mathbf{x}$, Eq.~(\ref{cwl2-eq4}) can be written as an unconstrained optimization problem. Moreover, Carlini et al. introduce the confidence parameter $\kappa$ into the above formulation. Above all, the CW-L2 attack seeks the adversarial image by solving the following problem

{
\begin{align}
\label{CWL2}
&\min_{\mathbf{w}} ||\frac{1}{2}\left(\tanh(\mathbf{w}) + 1\right) - \mathbf{x} ||_2^2 + c\cdot\\ \nonumber
&\max\left\{-\kappa, \max_{i\neq t}(Z(\frac{1}{2}(\tanh(\mathbf{w}))+1)_i) 
- Z(\frac{1}{2}(\tanh(\mathbf{w}))+1)_t \right\}.
\end{align}
}

The Adam optimizer \cite{Kingma:2014Adam} can solve this unconstrained optimization problem efficiently. All three attacks clip the values of the adversarial image $\mathbf{x}'$ to between 0 and 1. 


\begin{figure}[!ht]
\centering
\begin{tabular}{cc}
\includegraphics[width=0.45\columnwidth]{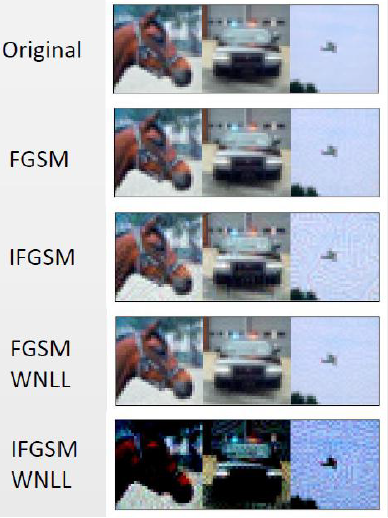}&
\includegraphics[width=0.45\columnwidth]{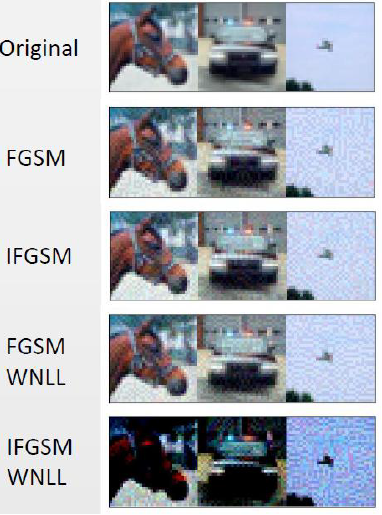}\\
(a)&(b)\\
\includegraphics[width=0.45\columnwidth]{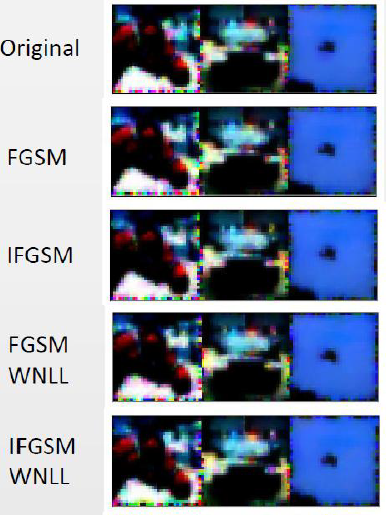}&
\includegraphics[width=0.45\columnwidth]{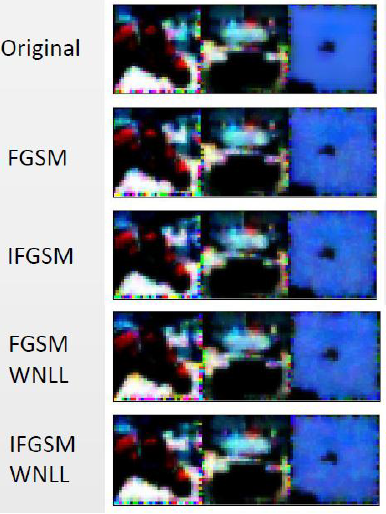}\\
(c)&(d)\\
\end{tabular}
\caption{Samples from CIFAR10. Panel (a): from the top to the last rows show the original, adversarial images by attacking ResNet56 with FGSM and IFGSM ($\epsilon=0.02$); and by attacking ResNet56-WNLL. Panel (b) corresponding to those in panel (a) with $\epsilon=0.08$. Charts (c) and (d) corresponding to the TV minimized images in (a) and (b), respectively.}
\label{fig:Adv-images}
\end{figure}

\subsection{Attack the DNN with WNLL Activation}
For a given mini-batch of testing images $(\mathbf{X}, \mathbf{Y})$ and template $(\mathbf{X}^{\rm te}, \mathbf{Y}^{\rm te})$, we denote the DNN-WNLL as $\tilde{\mathbf{Y}} = {\rm WNLL}(Z(\{\mathbf{X}, \mathbf{X}^{\rm te}\} ), \mathbf{Y}^{\rm te})$, where $Z(\{\mathbf{X}, \mathbf{X}^{\rm te}\} )$ is the composition of the DNN and buffer blocks as shown in Fig.~\ref{fig:WNLL-DNN-Structure}(c). By ignoring dependence of the loss function on the parameters, the loss function for DNN-WNLL can be written as $\tilde{\mathcal{L}}(\mathbf{X}, \mathbf{Y}, \mathbf{X}^{\rm te}, \mathbf{Y}^{\rm te}) \doteq {\rm Loss}(\hat{\mathbf{Y}}, \mathbf{Y})$. The above attacks for DNN-WNLL are formulated below.

\begin{itemize}
    \item {\bf FGSM} 
    
    \begin{align}
    \label{FGSM-WNLL}
    \mathbf{X}' = \mathbf{X} + \epsilon \cdot {\rm sign}\left( \nabla_\mathbf{X} \tilde{\mathcal{L}}(\mathbf{X}, \mathbf{Y}, \mathbf{X}^{\rm te}, \mathbf{Y}^{\rm te}) \right).
    \end{align}
    
    \item {\bf IFGSM} 
    \begin{align}
    \label{IFGSM-WNLL}
    \mathbf{X}^{(m)} = {\rm Clip}_{\mathbf{X}, \epsilon}[\mathbf{X}^{(m-1)} + 
     \alpha \cdot {\rm sign} \left( \nabla_{\mathbf{X}} \tilde{\mathcal{L}}(\mathbf{X}^{(m-1)}, \mathbf{Y}, \mathbf{X}^{\rm te}, \mathbf{Y}^{\rm te}) \right)],
    \end{align}
    where $m=1, 2, \cdots, M$;  $\mathbf{X}^{(0)} = \mathbf{X}$ and $\mathbf{X}'=\mathbf{X}^{(M)}$.
    
    \item {\bf CW-L2} 
    {
    \begin{align}\label{CWL2-WNLL}
    &\min_{\mathbf{W}}  ||\frac{1}{2}\left(\tanh(\mathbf{W}) + 1\right) - \mathbf{X} ||_2^2  + c\cdot  \max[-\kappa, \\ \nonumber
     &  \max_{\mathbf{i}\neq \mathbf{t} }( Z(\frac{1}{2}(\tanh(\mathbf{W}))+1)_\mathbf{i}) -  Z(\frac{1}{2}(\tanh(\mathbf{W}))+1)_\mathbf{t} ],
    \end{align}
    }
    where $\mathbf{i}$ are the logit values of the input images $\mathbf{X}$, $\mathbf{t}$ are the target labels. 
\end{itemize}

In the above attacks, $\nabla_{\mathbf{X}}\tilde{\mathcal{L}}$ is required to generate the adversarial images. In the DNN-WNLL, this gradient is difficult to compute. As shown in Fig.~\ref{fig:WNLL-DNN-Structure} (c), we approximate $\nabla_{\mathbf{X}}\tilde{\mathcal{L}}$ in the following way
{
\begin{eqnarray}
\label{GradientApprox}
\nabla_{\mathbf{X}}\tilde{\mathcal{L}} = \frac{\partial \mathcal{L}^{\rm WNLL}}{\partial\hat{\mathbf{Y}}}\cdot \frac{\partial\hat{\mathbf{Y}}}{\partial\hat{\mathbf{X}}} \cdot \frac{\partial\hat{\mathbf{X}}}{\partial\tilde{\mathbf{X}}} \cdot \frac{\partial\tilde{\mathbf{X}}}{\partial\mathbf{X}} 
\approx
\frac{\partial \mathcal{L}^{\rm Linear}}{\partial\tilde{\mathbf{Y}}}\cdot \frac{\partial\tilde{\mathbf{Y}}}{\partial\hat{\mathbf{X}}} \cdot \frac{\partial\hat{\mathbf{X}}}{\partial\tilde{\mathbf{X}}} \cdot \frac{\partial\tilde{\mathbf{X}}}{\partial\mathbf{X}},
\end{eqnarray}
}
again, in the above approximation, we set the value of $\mathcal{L}^{\rm Linear}$ to the value of $\tilde{\mathcal{L}}$.

Based on our numerical experiments, the batch size of $\mathbf{X}$ has minimal influence on the adversarial attack and defense. In all of our experiments, we choose the size of both mini-batches $\mathbf{X}$ and the template to be $500$.

\section{Defense by Interpolating Function, TVM, and Training Data Augmentation}\label{section:Interpolate:TVM}
To defend against adversarial attacks, we first combine the data-dependent activation with input transformation and with training data augmentation. We train ResNet56 \cite{ResNet} and ResNet56-WNLL, respectively, on the original training data, the TV minimized training data, and a combination of the previous two. 
Moreover, in testing, we apply the TVM \cite{ROF:1992} used by \cite{ChuanGuo:2018}, with the same setting, to transform the adversarial images to boost classification performance. The basic idea of TVM is to reconstruct the simplest image $\mathbf{z}$ from the sub-sampled image, $X\odot \mathbf{x}$ with $X$ the mask filled by a Bernoulli binary random variable, by solving
\begin{equation*}
\min_\mathbf{z}||(1-X)\odot (\mathbf{z} - \mathbf{x}) ||_2 + \lambda_{TV} \cdot TV_2(\mathbf{z}),    
\end{equation*}
where $\lambda_{TV} > 0$ is the regularization constant.

We apply the three attack schemes mentioned above to attack ResNet56 and ResNet56-WNLL. For IFGSM, we run 10 iterations of Eqs.~(\ref{IFGSM-2}) and (\ref{IFGSM-WNLL}) with $\epsilon=0.1$ to attack the DNN with two different output activations, respectively. For the CW-L2 attack (Eqs.~(\ref{CWL2}, \ref{CWL2-WNLL})), in both scenarios we set the parameters $c=10$ and $\kappa=0$, and run 10 iterations of the Adam optimizer with learning rate 0.01. 
Figure~\ref{fig:Adv-images} depicts three randomly selected images (horse, automobile, airplane) from the CIFAR10 dataset, as well as the perturbed images from applying different attacks on ResNet56 and ResNet56-WNLL, and the TV minimized ones. All attacks successfully fool the classifiers to classify any of them correctly. Figure~\ref{fig:Adv-images} (a) shows that the perturbations resulted from FGSM attack with $\epsilon=0.02$ is almost imperceptible. 
However, both FGSM and IFGSM attacks are powerful in fooling DNNs. Figure~\ref{fig:Adv-images} (b) shows the corresponding images of (a) with a stronger attack, $\epsilon=0.08$. With a larger $\epsilon$, the adversarial images become more noisy. The TV minimized images of Fig.~\ref{fig:Adv-images} (a) and (b) are shown in Fig.~\ref{fig:Adv-images} (c) and (d), respectively. TVM removes a significant amount of information from the original and the adversarial images. Meanwhile, it also makes it harder for humans to classify them.

\subsection{Numerical Results}
In this subsection, we first discuss the transferability of adversarial examples generated by attacking DNNs with softmax and WNLL activation functions. The transferability of adversarial examples is often used for black-box adversarial attacks. Adversarial examples of a robust DNN typically have good transferability. Next, we numerically verify the efficacy of adversarial defense by leveraging DNN with the WNLL activation function and TVM. Finally, we explain the adversarial robustness by considering the deep learning features learned by DNN with different activation functions.

\subsubsection{Transferability of the Adversarial Images}
Consider the transferability of adversarial examples crafted by using the above adversarial attacks to attack ResNet56 with either softmax or WNLL activation. We utilize the training strategy used in \cite{BaoWang:2018NIPS} to train the DNNs. To test the transferability, we classify the adversarial images by using ResNet56 with the opponent activation (the opponent activation of WNLL is softmax, and vice versa). We list the mutual classification accuracy (the accuracy of DNN with one specific activation to classify adversarial images crafted by attacking DNN with the other activation) on adversarial images resulting from using FGSM or IFGSM in Table.~\ref{Table1:Mutual-Classification}. The adversarial images crafted by attacking ResNet56 with two types of activation functions are both transferable, as the mutual classification accuracy on adversarial images ($\epsilon\neq 0$) is significantly lower than testing on the clean images ($\epsilon=0$). For both FGSM and IFGSM, the stronger attack (in the sense of bigger $\epsilon$) is adapted to the opponent activation function, as the mutual classification accuracy decreases dramatically as $\epsilon$ increases. IFGSM not only fools its underlying model  completely, but also significantly decreases the accuracy of the opponent DNN. The mutual classification results for the CW-L2 attack is shown in Table.~\ref{Table1:Mutual-Classification-WNLL}, where Exp-I denotes classifying adversarial images resulted from attacking ResNet56-WNLL by ResNet56, and Exp-II denotes the opposite. Training data augmentation can defend CW-L2 attack very effectively.

\begin{table}[!ht]
\centering
\fontsize{10.0}{10}\selectfont
\begin{threeparttable}
\caption{Mutual classification accuracy on the adversarial images crafted by using FGSM and IFGSM to attack ResNet56 and ResNet56-WNLL. (Unit: $\%$)}\label{Table1:Mutual-Classification}
\begin{tabular}{cccccccc}
\toprule
Attack    &Training data & $\epsilon=0$    &   $\epsilon=0.02$ &  $\epsilon=0.04$ &  $\epsilon=0.06$ &  $\epsilon=0.08$ &  $\epsilon=0.1$ \cr
\midrule
\multicolumn{8}{c}{Accuracy of ResNet56 on adversarial images crafted by attacking ResNet56-WNLL}\cr
\midrule
FGSM              & Original data       &93.0    &69.8      &56.9   &44.6  &34.6     &28.3 \cr
FGSM              & TVM data            &88.3    &51.5      &37.9   &30.1  &24.7     &20.9 \cr
FGSM              & Original + TVM &93.1    &78.5      &70.9   &64.6  &59.8     &55.8 \cr
\midrule
IFGSM              & Original data      &93.0    &5.22      &5.73   &6.73  &7.55     &8.55 \cr
IFGSM              & TVM data           &88.3    &7.00      &6.82   &8.30  &9.28     &10.7 \cr
IFGSM              & Original + TVM&93.1    &27.3      &28.6   &29.5  &29.1     &29.4 \cr
\toprule
\multicolumn{7}{c}{Accuracy of ResNet56-WNLL on adversarial images crafted by attacking ResNet56}\cr
\midrule
FGSM              & Original data       &94.5    &65.2      &49.0   &39.3  &32.8     &28.3 \cr
FGSM              & TVM data            &90.6    &45.9      &30.9   &22.2  &16.9     &13.8 \cr
FGSM              & Original + TVM data &94.7    &78.3      &68.2   &61.1  &56.5     &52.5 \cr
\midrule
IFGSM              & Original data      &94.5    &3.37      &3.71   &3.54  &4.69     &6.41 \cr
IFGSM              & TVM data           &90.6    &7.88      &7.51   &7.58  &8.07     &9.67 \cr
IFGSM              & Original + TVM data&94.7    &34.3      &33.4   &33.1  &34.6     &35.8 \cr
\bottomrule
\end{tabular}
\end{threeparttable}
\end{table}

\begin{table}[!ht]
\centering
\fontsize{10.0}{10}\selectfont
\begin{threeparttable}
\caption{Mutual classification accuracy on the adversarial images crafted by using CW-L2 to attack ResNet56 and ResNet56-WNLL. (Unit: $\%$)}\label{Table1:Mutual-Classification-WNLL}
\begin{tabular}{cccc}
\toprule
\ \ \ \ \ Training data \ \ \ \    &\ \ \ \  Original data\ \ \ \  & \ \ \ \ TVM data\ \ \ \ &  \ \ \  Original + TVM data \cr
\midrule
 Exp-I  &   52.1  & 43.2  & 80.0\\
\midrule
 Exp-II &  59.7  & 41.1   & 80.1 \\
\bottomrule
\end{tabular}
\end{threeparttable}
\end{table}

\begin{table}[!ht]
\centering
\fontsize{10.0}{10}\selectfont
\begin{threeparttable}
\caption{Testing accuracy on the adversarial/TVM adversarial CIFAR10 dataset. The testing accuracy with no defense is in red italic; and the results with all three defenses are in boldface. (Unit: $\%$)}\label{Table1:adversarialdat-WNLL}
\begin{tabular}{cccc}
\toprule
\ \ \ \ \ Training data\ \ \ \ \     &\ \ \  Original data\ \ \  &\ \ \   TVM data\ \ \  & \ \ \  Original + TVM data\ \ \  \cr
\midrule
ResNet56  &   \textcolor{red}{\it 4.94}/32.2  & 11.8/54.0  & 15.1/52.4\\
\midrule
ResNet56-WNLL &  18.3/35.2  & 15.0/53.9   & 28/{\bf 54.5} \\
\bottomrule
\end{tabular}
\end{threeparttable}
\end{table}

\begin{table}[!ht]
\centering
\fontsize{10.0}{10}\selectfont
\begin{threeparttable}
\caption{Testing accuracy on the adversarial/TVM adversarial CIFAR10 dataset. The testing accuracy with no defense is in red italic; and the results with all three defenses are in boldface. (Unit: $\%$)}\label{Table1:adversarialdata}
\begin{tabular}{cccccccc}
\toprule
Attack     &Training data    &  $\epsilon=0$ &  $\epsilon=0.02$ &  $\epsilon=0.04$ &  $\epsilon=0.06$ &  $\epsilon=0.08$ &  $\epsilon=0.1$ \cr
\midrule
\multicolumn{8}{c}{ResNet56}\cr
\midrule
FGSM              & Original data        &93.0   &\textcolor{red}{\it 36.9}/19.4      &\textcolor{red}{\it 29.6}/18.9  &\textcolor{red}{\it 26.1}/18.4  &\textcolor{red}{\it 23.1}/17.9 &\textcolor{red}{\it 20.5}/17.1 \cr
FGSM              & TVM data             &88.3   &27.4/50.4      &19.1/47.2   &16.6/43.7  &15.0/38.9 &13.7/35.0 \cr
FGSM              & Original + TVM  &93.1   &48.6/51.1      &42.0/47.6   &39.1/44.2  &37.1/41.8 &35.6/39.1 \cr
\midrule
IFGSM              & Original data       &93.0   &\textcolor{red}{\it 0}/16.6      &\textcolor{red}{\it 0}/16.1   &\textcolor{red}{\it 0.02}/15.9   &\textcolor{red}{\it 0.1}/15.5 &\textcolor{red}{\it 0.25}/16.1 \cr
IFGSM              & TVM data            &88.3   &0.01/43.4      &0/42.5    &0.02/42.4   &0.18/42.7 &0.49/42.4 \cr
IFGSM              & Original + TVM &93.1   &0.1/38.4      &0.09/37.9   &0.36/37.9  &0.84/37.6&1.04/37.9 \cr
\toprule
\multicolumn{8}{c}{ResNet56-WNLL}\cr
\midrule
FGSM              & Original data        &94.5   &58.5/26.0      &50.1/25.4   &42.3/25.5  &35.7/24.9 &29.2/22.9 \cr
FGSM              & TVM data             &90.6   &31.5/52.6      &24.5/49.6   &20.2/45.3  &17.3/41.6 &14.4/37.5 \cr
FGSM              & Original + TVM  &94.7   &{\bf60.5}/ 55.4      &{\bf56.7}/52.0   &{\bf55.3}/48.6  &{\bf53.2}/45.9 &{\bf50.1}/43.7 \cr
\midrule
IFGSM              & Original data       &94.5   &0.49/16.7      &0.14/17.3   &0.3/16.9  &1.01/16.6 &0.94/16.5 \cr
IFGSM              & TVM data            &90.6   &0.61/37.3      &0.43/36.3   &0.63/35.9 &0.87/35.9 &1.19/35.5 \cr
IFGSM              & Original + TVM &94.7   &0.19/{\bf38.5 }     &0.3/{\bf39.4 }  &0.63/{\bf 40.1}  &1.26/{\bf 38.9} &1.72/{\bf 39.1} \cr
\bottomrule
\end{tabular}
\end{threeparttable}
\end{table}

\subsubsection{Adversarial Defense}
\begin{figure}[!ht]
\centering
\begin{tabular}{cc}
\includegraphics[width=0.46\columnwidth]{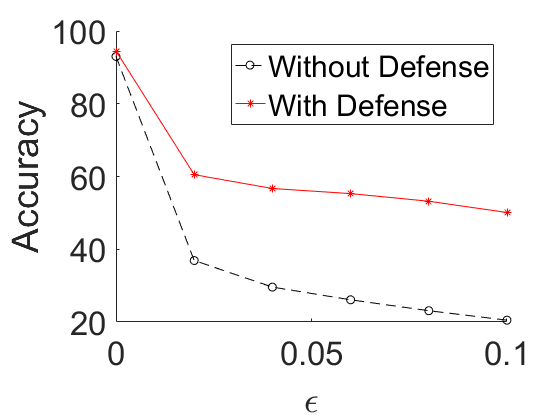}&
\includegraphics[width=0.46\columnwidth]{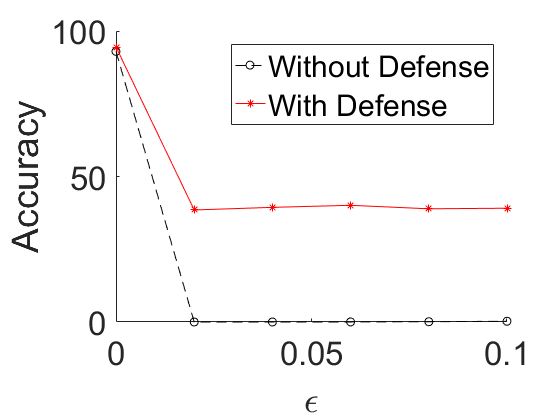}\\
(a)&(b)\\
\end{tabular}
\caption{$\epsilon$ v.s. accuracy without defense, and defending by WNLL activation, TVM and augmented training. (a) and (b) plot results for FGSM and IFGSM attack, respectively.}
\label{fig:Defense-Res}
\end{figure}

Figure~\ref{fig:Defense-Res} plots the results of adversarial defense by combining the WNLL activation, TVM, and training data augmentation. Panels (a) and (b) show the testing accuracy of ResNet56 with and without defense on CIFAR10 data for FGSM and IFGSM, respectively. It is seen that as $\epsilon$ increases, the testing accuracy decreases rapidly. FGSM is a relatively weak attack, and the accuracy remains above 20.5$\%$ even with the most potent attack ($\epsilon=0.1$). Meanwhile, the defense raises the accuracy to 43.7$\%$. Figure~\ref{fig:Defense-Res} (b) shows that IFGSM fools ResNet56 near completely even with $\epsilon=0.02$. The defense maintains the accuracy above 38.5$\%$, 54.5$\%$ under the CW-L2 and IFGSM attacks, respectively (see Tables.~\ref{Table1:adversarialdat-WNLL} and \ref{Table1:adversarialdata}). Compared to the state-of-the-art defensive methods on CIFAR10, PixelDefend, our approach is much simpler and faster. Without adversarial training, we have shown our defense is more robust to FGSM and IFGSM attacks under the strongest attack than PixelDefend \cite{Song:2018}. Moreover, our defense strategy is additive to adversarial training and many other defenses including PixelDefend.

To analyze the contribution from each component of the defensive strategy, we separate the three parts and list the testing accuracy in Tables.~\ref{Table1:adversarialdat-WNLL} and \ref{Table1:adversarialdata}. Performing TVM on the adversarial images cannot defend FGSM attacks except when the training data contains the TV minimized images. For instance, when we attack the model by FGSM with $\epsilon=0.02$, the accuracy on the adversarial images for ResNet56 and ResNet56-WNLL are 36.9$\%$ and 58.5$\%$, respectively, provided the models are trained on the original training data. The accuracy reduces to 19.4$\%$ and 26.0$\%$ when testing on the TV minimized adversarial images. For ResNet56, the accuracy raises to 50.4$\%$ and 51.1$\%$ when the model is trained on the TVM and augmented data, respectively. For ResNet-WNLL, the accuracy increases to 52.6$\%$ and 55.4$\%$, respectively.
The WNLL activation improves testing accuracy of adversarial attacks significantly and persistently. Augmented training can also improve the stability consistently.

IFGSM fools the ResNet56-WNLL near completely, as the accuracy is always less than or close to 1$\%$. These results verify the efficacy of using the approximated gradient, i.e., Eq.~(\ref{GradientApprox}), in attacking the neural nets.

\begin{algorithm}[!ht]
\caption{PGD Adversarial Training of the DNN-WNLL}\label{PGD-WNLL}
\begin{algorithmic}
\State \textbf{Input: } Training set: (data, label) pairs $(\mathbf{X}, \mathbf{Y})$. $N_1$:  the number of epochs in training DNN $+$ Linear blocks. $N_{\rm IFGSM}$: the number of iterations of IFGSM attack.
\State \textbf{Output: } An optimized DNN-WNLL, denoted as ${\rm DNN}_{\rm WNLL}$.
\For {${\rm iter} = 1, \dots, N$ (where $N$ is the number of alternating steps.)}
\State //{\it PGD adversarial training of the left branch: DNN with linear activation}.
\State Train DNN $+$ Linear blocks, and denote the learned model as ${\rm DNN}_{\rm Linear}$.
\State Partition the training data into $M_1$ mini-batches, i.e., $(\mathbf{X}, \mathbf{Y}) = \bigcup_{i=1}^{M_1} (\mathbf{X}_i, \mathbf{Y}_i)$.
\For{${\rm epoch_1} = 1, $\dots$, N_1$}
\For{$i = 1, $\dots$, M_1$}
\State //{\it Attack the input images by IFGSM.}
\For{${\rm iter_1} = 1, \dots,  N_{\rm \footnotesize IFGSM}$}
\State Update the training image $\mathbf{X}_i = \mathbf{X}_i + \epsilon\cdot {\rm sign}\left(\nabla_{\mathbf{X}_i} \mathcal{L} \right)$, \\
\hskip 2.1cm where $\mathcal{L}$ is the loss of the prediction by DNN $+$ Linear blocks w.r.t\\
\hskip 2.1cm the ground truth labels $\mathbf{Y}_i$.
\EndFor
\State Backpropagate the classification error of the adversarial images.
\EndFor
\EndFor

\State //{\it PGD adversarial training of the right branch: DNN with WNLL activation.}
\State Split $(\mathbf{X}, \mathbf{Y})$ into training data and template, i.e.,\\
\hskip 1.1cm $(\mathbf{X}, \mathbf{Y}) \doteq (\mathbf{X}^{\rm tr}, \mathbf{Y}^{\rm tr}) \bigcup (\mathbf{X}^{\rm te}, \mathbf{Y}^{\rm te})$.
\State Partition the training data into $M_2$ mini-batches, i.e.,\\ 
\hskip 1.1cm $(\mathbf{X}^{\rm tr}, \mathbf{Y}^{\rm tr}) = \bigcup_{i=1}^{M_2} (\mathbf{X}_i^{\rm tr}, \mathbf{Y}_i^{\rm tr})$.
\For{${\rm epoch_2} = 1, $\dots$, N_2$}
\For{$i=1, \dots, M_2$}
\State //{\it Attack the input training images by IFGSM.}
\For{${\rm iter_1} = 1, \dots,  N_{\rm \footnotesize IFGSM}$}
\State Update the training image $\mathbf{X}^{\rm tr}_i = \mathbf{X}^{\rm tr}_i + \epsilon\cdot {\rm sign}\left(\nabla_{\mathbf{X}^{\rm tr}_i} \tilde{\mathcal{L}} \right)$, \\
\hskip 2.1cm where $\tilde{\mathcal{L}}$ is the loss of the prediction by DNN with WNLL\\
\hskip 2.1cm activation w.r.t the ground truth labels $\mathbf{Y}^{\rm tr}_i$.\\
\EndFor
\State Backpropagate the classification error of the adversarial images.
\EndFor
\EndFor
\EndFor
\end{algorithmic}
\end{algorithm}

\subsubsection{Analysis of the Geometry of Features} 
\begin{figure}[!ht]
\centering
\begin{tabular}{cc}
\includegraphics[width=0.46\columnwidth]{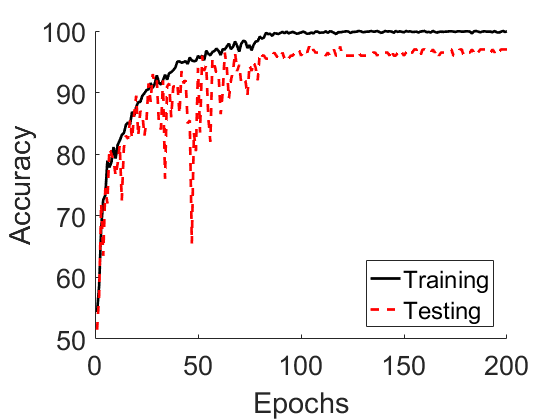} &
\includegraphics[width=0.46\columnwidth]{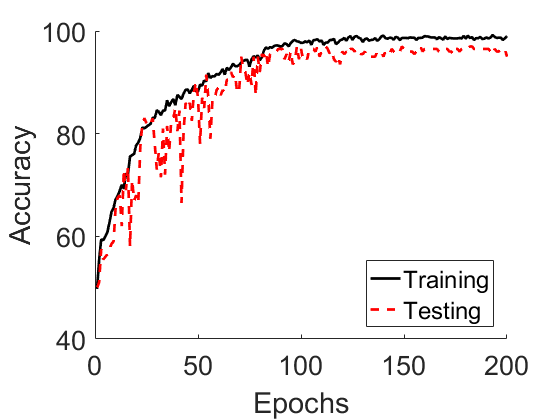}\\
(a)&(b)\\
\end{tabular}
\caption{Epochs v.s. accuracy of ResNet56 on CIFAR10. (a): without the additional FC layer; (b): with the additional FC layer.}
\label{fig:Add2x2}
\end{figure}

We consider features' geometry of the original and adversarial images.
We randomly select 1000 training and 100 testing images from the airplane and automobile classes, respectively.
We apply two visualization strategies for ResNet56: (1) Apply the principle component analysis (PCA) to reduce the 64D features from the layer before the softmax to 2D,
and (2) we add a 2 by 2 fully connected (FC) layer before the softmax to learn 2D features.
We verify that the newly added layer does not change the performance of ResNet56, as shown in Fig.~\ref{fig:Add2x2}, and the training and testing performance remains essentially the same.

\begin{figure}[!ht]
\centering
\begin{tabular}{cccc}
\includegraphics[width=0.22\columnwidth]{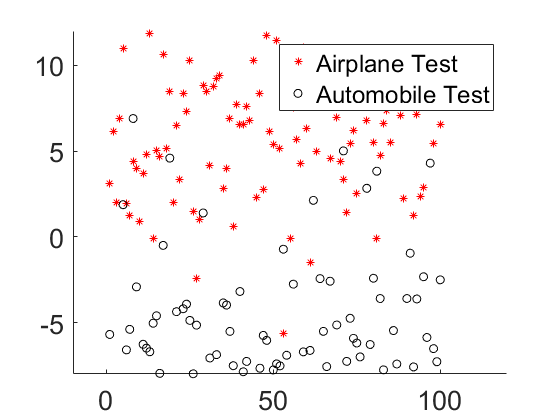}&
\includegraphics[width=0.22\columnwidth]{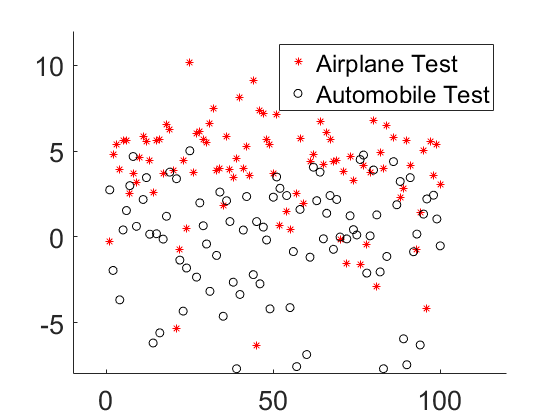}&
\includegraphics[width=0.22\columnwidth]{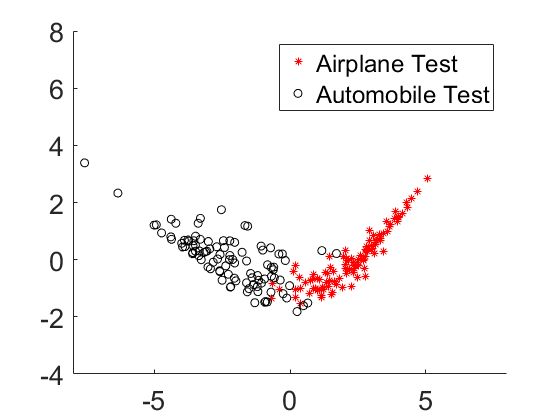}&
\includegraphics[width=0.22\columnwidth]{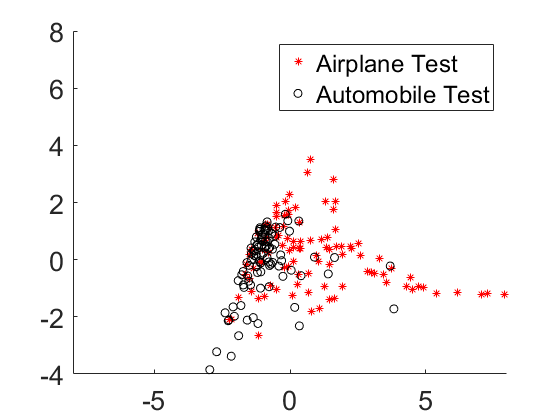}\\
(a)&(b)&(c)&(d)\\
\includegraphics[width=0.22\columnwidth]{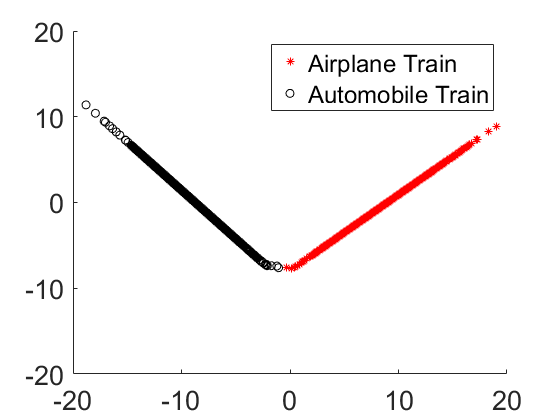}&
\includegraphics[width=0.22\columnwidth]{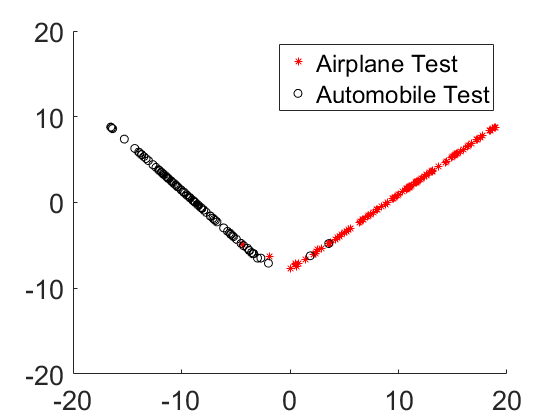}&
\includegraphics[width=0.22\columnwidth]{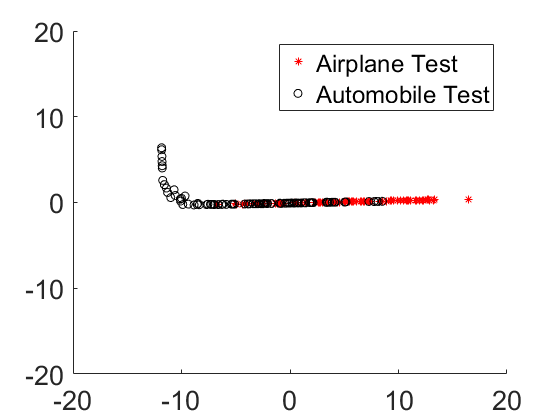}&
\includegraphics[width=0.22\columnwidth]{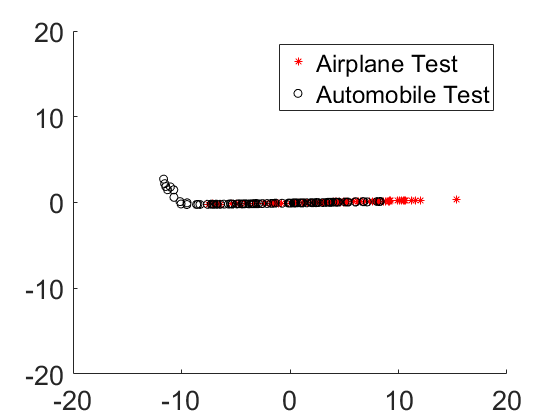}\\
(e)&(f)&(g)&(h)\\
\end{tabular}
\caption{Visualization of the features learned by DNN with softmax ((a), (b), (c), (d)) and WNLL ((e), (f), (g), (h)) activation functions. (a) and (b) plot the 2D features of the original and adversarial testing images; (c) and (d) are the first two principle components of the 64D features for the original and adversarial testing images, respectively. Charts (e), (f) plot the first two components of the training and testing features learned by ResNet56-WNLL; (g) and (h) show the two principle components of the adversarial and TV minimized adversarial images for the test set.
}
\label{fig:FeaturesGeometry}
\end{figure}

Figure~\ref{fig:FeaturesGeometry} (a) and (b) show the 2D features generated by ResNet56 with the additional FC layer for the original and adversarial testing images, respectively, where we generate the adversarial images by using FGSM ($\epsilon=0.02$). Before adversarial perturbation (Fig.~\ref{fig:FeaturesGeometry} (a)), there is a line that can separate the two classes very well. The small perturbation mixes the features and there is no linear classifier that can easily separate these two classes (Fig.~\ref{fig:FeaturesGeometry} (b)). The first two principle components (PCs) of the 64D features of the clean and adversarial images are shown in Fig.~\ref{fig:FeaturesGeometry} (c) and (d), respectively. Again, the PCs are well separated for clean images, while adversarial images causes overlap.

The bottom charts of Fig.~\ref{fig:FeaturesGeometry} depict the first two PCs of the 64D features output from the layer before the WNLL. The distributions of the unperturbed training and testing data are the same, as illustrated in panels (e) and (f). The new features are better separated which indicates that DNN-WNLL are more accurate and more robust to small random perturbation. Panels (g) and (h) plot the features of the adversarial and TV minimized adversarial images in the test set. The adversarial attacks make the features move towards each other and TVM helps to eliminate the outliers. Based on our computation, the interpolating function on features shown in panels (g) and (h) are significantly more accurate than the softmax classifier as shown in panel (d).
The fact that the adversarial perturbations change the features' distribution was also noticed in \cite{Song:2018}, and \cite{Liao_2018_CVPR}. 

\section{PGD Adversarial Training with Data-Dependent Activation Function}\label{section:PGD}
Image transformation based adversarial defense has been broken recently by circumventing the obfuscated gradient \cite{Athalye:2018}. To train a DNN that is most resistant to adversarial attacks, Madry et al. \cite{Madry:2018}, incorporate the adversarial perturbation into the empirical risk function $\mathbb{E}_{(\mathbf{x}, y)\sim \mathcal{D}}\left[\mathcal{L}(\mathbf{x}, y, \theta)\right]$, where $\mathcal{D}$ 
is the collection of the pairs of training images and labels, and $\theta$ represents the parameters of the neural nets. The idea of PGD adversarial training is that instead of feeding samples from $\mathcal{D}$ directly into the loss $\mathcal{L}$, we use the adversary to perturb the input first, and then we end up with the following saddle point problem
\begin{equation}
\label{saddle-PGD}
\min_\theta \rho(\theta) = \min_\theta \mathbb{E}_{(\mathbf{x}, y)\sim \mathcal{D}}\left[\max_{\delta\in S}\mathcal{L}(\theta, \mathbf{x}+\delta, y)\right],
\end{equation}
where $\delta$ is the adversarial perturbation. To make the problem (Eq.~(\ref{saddle-PGD})) solvable, the inner maximization problem is relaxed to a strong adversarial attack, say IFGSM. It is argued in \cite{Athalye:2018}, that PGD adversarial training achieves the best resistance to adversarial attacks for CIFAR10 classification. We extend the PGD adversarial training to DNN-WNLL by applying the approximated gradient, Eq.~(\ref{GradientApprox}), to approximately resolve the interior maximization problem. We summarize the PGD adversarial training of DNN-WNLL in Algorithm~\ref{PGD-WNLL}.

\subsection{Numerical Results}
We consider PGD adversarial training, respectively, for the ResNet20 and ResNet20-WNLL. Again, we train the ResNet20 with two types of activation, where we follow the strategy used in \cite{BaoWang:2018NIPS}, and where all the hyper-parameters in Algorithm~\ref{PGD-WNLL} are referred. To approximate $\max_{\delta\in S}\mathcal{L}(\theta, \mathbf{x}+\delta, y)$, we apply the IFGSM attack with $\alpha=8/255$ in Eqs.~(\ref{IFGSM-2}, \ref{IFGSM-WNLL}).

\begin{figure}[!ht]
\centering
\begin{tabular}{cc}
\includegraphics[width=0.45\columnwidth]{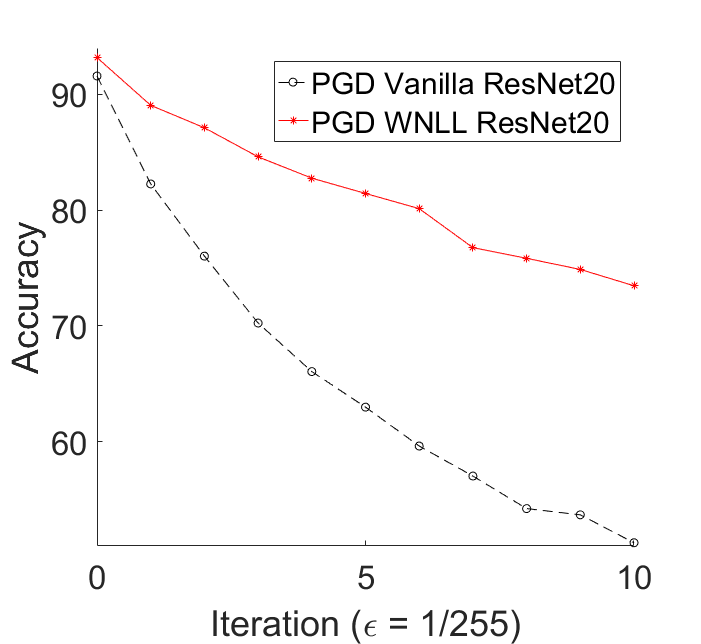}&
\includegraphics[width=0.45\columnwidth]{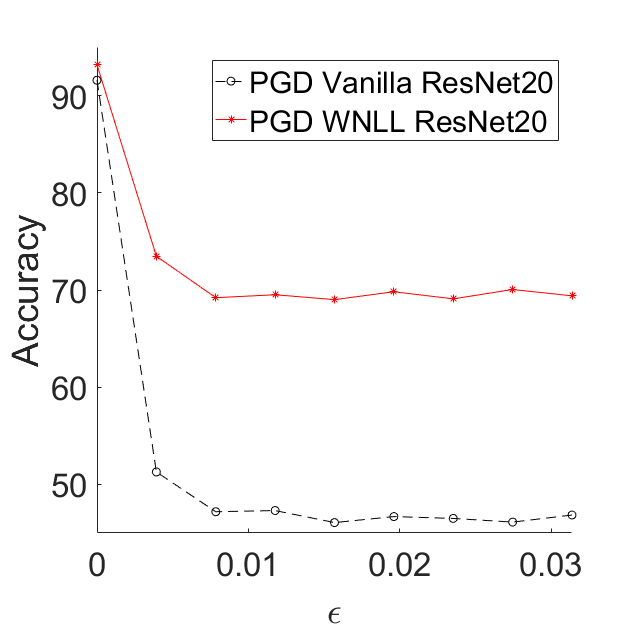}\\
(a) & (b)\\
\end{tabular}
\caption{(a): $\#$IFGSM iterations v.s. accuracy for the ResNet20 and the ResNet20-WNLL trained with PGD adversarial training. (b):$\epsilon$ v.s. accuracy for the ResNet20 and the ResNet20-WNLL trained with PGD adversarial training.}
\label{fig:Defense-Res2}
\end{figure}

First, we fixed the attack strength $\epsilon=1/255$ and vary the number of IFGSM iterations. As shown in Fig.~\ref{fig:Defense-Res2} (a), the accuracy of ResNet20 with both activations decreases as the number of iteration increases. The vanilla ResNet20's accuracy decays much faster than the ResNet20-WNLL. The difference is $\sim 23\%$ when 10 iterations of IFGSM is applied. Second, we fixed the IFGSM iteration to be 10 and vary $\epsilon$ from $0$ to $8/255$ with step size $1/255$. As shown in Fig.~\ref{fig:Defense-Res2} (b), for different nonzero attack strengths, PGD adversarial training of the ResNet20-WNLL has $\sim 23\%$ higher accuracy than the vanilla one consistently.

\section{Concluding Remarks}
In this paper, by analyzing the influence of adversarial perturbations on the geometric structure of the DNN features, we propose to defend against adversarial attacks by using a data-dependent activation function. We further show our defenses are additive to other defenses, namely total variation minimization, training data augmentation, and projected gradient descent adversarial training. Results on ResNet20 and ResNet56 with CIFAR10 benchmark reveal that these defenses improve robustness to adversarial perturbation significantly. Total variation minimization simplifies the adversarial images, which is very useful in removing adversarial perturbation. The data-dependent activation framework raises the accuracy of PGD adversarial training around $23\%$ under different attack strengths. An interesting direction to explore is to combine these methods with other denoising methods to remove adversarial perturbation.

\section*{Acknowledgments} This material is based on research sponsored by the National Science Foundation under grant number DMS-1924935 and DMS-1554564 (STROBE). The Air Force Research Laboratory under grant numbers FA9550-18-0167 and MURI FA9550-18-1-0502, the Office of Naval Research under grant number N00014-18-1-2527. ALB is partially supported by the Simons Math $+$ X award.


\begin{thebibliography}{99}

\bibitem{Akhtar:2018}
N.~Akhtar and A.~Mian.
\newblock Threat of adversarial attacks on deep learning in computer vision: A
  survey.
\newblock {\em arXiv preprint arXiv:1801.00553}, 2018.

\bibitem{Akhtar_2018_CVPR}
Naveed Akhtar, Jian Liu, and Ajmal Mian.
\newblock Defense against universal adversarial perturbations.
\newblock In {\em The IEEE Conference on Computer Vision and Pattern
  Recognition (CVPR)}, June 2018.

\bibitem{Athalye:2018}
A.~Athalye, N.~Carlini, and D.~Wagner.
\newblock Obfuscated gradients give a false sense of security: Circumventing
  defenses to adversarial examples.
\newblock {\em International Conference on Machine Learning}, 2018.

\bibitem{Athalye:2018B}
A.~Athalye, L.~Engstrom, A.~Ilyas, and K.~Kwok.
\newblock Synthesizing robust adversarial examples.
\newblock {\em International Conference on Machine Learning}, 2018.

\bibitem{BengioStraightTE:2013}
Y.~Bengio, N.~Leonard, and A.~Courville.
\newblock Estimating or propagating gradients through stochastic neurons for
  conditional computation.
\newblock {\em arXiv preprint arXiv:1308.3432}, 2013.

\bibitem{Brendel:2017}
W.~Brendel, J.~Rauber, and M.~Bethge.
\newblock Decision-based adversarial attacks: Reliable attacks against
  black-box machine learning models.
\newblock {\em arXiv preprint arXiv:1712.04248}, 2017.

\bibitem{CWAttack:2016}
N.~Carlini and D.A. Wagner.
\newblock Towards evaluating the robustness of neural networks.
\newblock {\em IEEE European Symposium on Security and Privacy}, pages 39--57,
  2016.

\bibitem{Dong_2018_CVPR}
Yinpeng Dong, Fangzhou Liao, Tianyu Pang, Hang Su, Jun Zhu, Xiaolin Hu, and
  Jianguo Li.
\newblock Boosting adversarial attacks with momentum.
\newblock In {\em The IEEE Conference on Computer Vision and Pattern
  Recognition (CVPR)}, June 2018.

\bibitem{Goodfellow:2014AdversarialTraining}
I.~J. Goodfellow, J.~Shlens, and C.~Szegedy.
\newblock Explaining and harnessing adversarial examples.
\newblock {\em arXiv preprint arXiv:1412.6275}, 2014.

\bibitem{PapernotMalware:2016}
K.~Grosse, N.~Papernot, P.~Manoharan, M.~Backes, and P.~McDaniel.
\newblock Adversarial perturbations against deep neural networks for malware
  classification.
\newblock {\em arXiv preprint arXiv:1606.04435}, 2016.

\bibitem{Giusti:2016Drones}
A.~Guisti, J.~Guzzi, D.C. Ciresan, F.L. He, J.P. Rodriguez, F.~Fontana,
  M.~Faessler, C.~Forster, J.~Schmidhuber, G.~Di Carlo, and et~al.
\newblock A machine learning approach to visual perception of forecast trails
  for mobile robots.
\newblock {\em IEEE Robotics and Automation Letters}, pages 661--667, 2016.

\bibitem{ChuanGuo:2018}
Chuan Guo, Mayank Rana, Moustapha Cisse, and Laurens van~der Maaten.
\newblock Countering adversarial images using input transformations.
\newblock In {\em International Conference on Learning Representations}, 2018.

\bibitem{ResNet}
K.~He, X.~Zhang, S.~Ren, and J.~Sun.
\newblock Deep residual learning for image recognition.
\newblock In {\em CVPR}, pages 770--778, 2016.

\bibitem{Ilyas:2018}
A.~Ilyas, L.~Engstrom, A.~Athalye, and J.~Lin.
\newblock Black-box adversarial attacks with limited queries and information.
\newblock {\em International Conference on Machine Learning}, 2018.

\bibitem{Kingma:2014Adam}
D.~Kingma and J.~Ba.
\newblock Adam: A method for stochastic optimization.
\newblock {\em arXiv preprint arXiv:1412.6980}, 2014.

\bibitem{Kurakin:2016}
A.~Kurakin, I.~J. Goodfellow, and S.~Bengio.
\newblock Adversarial examples in the physical world.
\newblock {\em arXiv preprint arXiv:1607.02533}, 2016.

\bibitem{2018arXiv180703888L}
K.~{Lee}, K.~{Lee}, H.~{Lee}, and J.~{Shin}.
\newblock {A Simple Unified Framework for Detecting Out-of-Distribution Samples
  and Adversarial Attacks}.
\newblock {\em ArXiv e-prints}, July 2018.

\bibitem{Liao_2018_CVPR}
Fangzhou Liao, Ming Liang, Yinpeng Dong, Tianyu Pang, Xiaolin Hu, and Jun Zhu.
\newblock Defense against adversarial attacks using high-level representation
  guided denoiser.
\newblock In {\em The IEEE Conference on Computer Vision and Pattern
  Recognition (CVPR)}, June 2018.

\bibitem{LiuYanpei:2016}
Y.~Liu, X.~Chen, C.~Liu, and D.~Song.
\newblock Delving into transferable adversarial examples and black-box attacks.
\newblock {\em arXiv preprint arXiv:1611.02770}, 2016.

\bibitem{DBLP:journals/corr/LuoBRPZ15}
Yan Luo, Xavier Boix, Gemma Roig, Tomaso~A. Poggio, and Qi~Zhao.
\newblock Foveation-based mechanisms alleviate adversarial examples.
\newblock {\em CoRR}, abs/1511.06292, 2015.

\bibitem{Madry:2018}
Aleksander Madry, Aleksandar Makelov, Ludwig Schmidt, Dimitris Tsipras, and
  Adrian Vladu.
\newblock Towards deep learning models resistant to adversarial attacks.
\newblock In {\em International Conference on Learning Representations}, 2018.

\bibitem{Moosavi-Dezfooli_2017_CVPR}
Seyed-Mohsen Moosavi-Dezfooli, Alhussein Fawzi, Omar Fawzi, and Pascal
  Frossard.
\newblock Universal adversarial perturbations.
\newblock In {\em The IEEE Conference on Computer Vision and Pattern
  Recognition (CVPR)}, July 2017.

\bibitem{DBLP:journals/corr/abs-1802-06806}
Seyed{-}Mohsen Moosavi{-}Dezfooli, Ashish Shrivastava, and Oncel Tuzel.
\newblock Divide, denoise, and defend against adversarial attacks.
\newblock {\em CoRR}, abs/1802.06806, 2018.

\bibitem{Na:2018}
Taesik Na, Jong~Hwan Ko, and Saibal Mukhopadhyay.
\newblock Cascade adversarial machine learning regularized with a unified
  embedding.
\newblock In {\em International Conference on Learning Representations}, 2018.

\bibitem{PapernotAttack:2016}
N.~Papernot, P.~McDaniel, S.~Jha, M.~Fredrikson, Z.B. Celik, and A.~Swami.
\newblock The limitations of deep learning in adversarial settings.
\newblock {\em IEEE European Symposium on Security and Privacy}, pages
  372--387, 2016.

\bibitem{PapernotSecurity:2016}
N.~Papernot, P.~McDaniel, A.~Sinha, and M.~Wellman.
\newblock Sok: Towards the science of security and privacy in machien learning.
\newblock {\em arXiv preprint arXiv:1611.03814}, 2016.

\bibitem{PapernotDistillationDefense:2016}
N.~Papernot, P.~McDaniel, X.~Wu, S.~Jha, and A.~Swami.
\newblock Distillation as a defense to adversarial perturbations against deep
  neural networks.
\newblock {\em IEEE European Symposium on Security and Privacy}, 2016.

\bibitem{DBLP:journals/corr/PapernotMG16}
Nicolas Papernot, Patrick~D. McDaniel, and Ian~J. Goodfellow.
\newblock Transferability in machine learning: from phenomena to black-box
  attacks using adversarial samples.
\newblock {\em CoRR}, abs/1605.07277, 2016.

\bibitem{Prakash_2018_CVPR}
Aaditya Prakash, Nick Moran, Solomon Garber, Antonella DiLillo, and James
  Storer.
\newblock Deflecting adversarial attacks with pixel deflection.
\newblock In {\em The IEEE Conference on Computer Vision and Pattern
  Recognition (CVPR)}, June 2018.

\bibitem{ROF:1992}
L.~Rudin, S.~Osher, and E.~Fatemi.
\newblock Nonlinear total variation based noise removal algorithms.
\newblock {\em Physica D: nonlinear phenomena}, pages 259--268, 1992.

\bibitem{Samangouei:2018}
Pouya Samangouei, Maya Kabkab, and Rama Chellappa.
\newblock Defense-{GAN}: Protecting classifiers against adversarial attacks
  using generative models.
\newblock In {\em International Conference on Learning Representations}, 2018.

\bibitem{WNLL:2018}
Z.~Shi, B.~Wang, and S.~J. Osher.
\newblock Error estimation of weighted nonlocal laplacian on random point
  cloud.
\newblock {\em arXiv preprint arXiv:1809.08622}, 2014.

\bibitem{Song:2018}
Yang Song, Taesup Kim, Sebastian Nowozin, Stefano Ermon, and Nate Kushman.
\newblock Pixeldefend: Leveraging generative models to understand and defend
  against adversarial examples.
\newblock In {\em International Conference on Learning Representations}, 2018.

\bibitem{Szegedy:2013}
C.~Szegedy, W.~Zaremba, I.~Sutskever, J.~Bruna, D.~Erhan, and I.~Goodfellow.
\newblock Intriguing properties of neural networks.
\newblock {\em arXiv preprint arXiv:1312.6199}, 2013.

\bibitem{tramer2018ensemble}
Florian Tramèr, Alexey Kurakin, Nicolas Papernot, Ian Goodfellow, Dan Boneh,
  and Patrick McDaniel.
\newblock Ensemble adversarial training: Attacks and defenses.
\newblock In {\em International Conference on Learning Representations}, 2018.


\bibitem{BaoWang:2018NIPS}
B.~Wang, X.~Luo, Z.~Li, W.~Zhu, Z.~Shi, and S.~Osher.
\newblock Deep neural nets with interpolating function as output activation.
\newblock {\em arXiv preprint arXiv:1802.00168}, 2018.


\bibitem{wang2019resnets}
Bao Wang, Zuoqiang Shi, and Stanley Osher.
\newblock ResNets Ensemble via the Feynman-Kac Formalism to Improve Natural and Robust Accuracies.
\newblock In {\em Advances in Neural Information Processing Systems}, 2019.

\bibitem{BaoWang:2019}
Bao Wang, and Stanley Osher.
\newblock Graph Interpolating Activation Improves Both Natural and Robust Accuracies in Data-Efficient Deep Learning.
\newblock {\em arXiv preprint arXiv:1907.06800}, 2019.


\bibitem{Wu:2018}
X.~Wu, U.~Jang, J.~Chen, L.~Chen, and S.~Jha.
\newblock Reinforcing adversarial robustness using model confidence induced by
  adversarial training.
\newblock {\em International Conference on Machine Learning}, 2018.

\bibitem{ie:2018}
Cihang Xie, Jianyu Wang, Zhishuai Zhang, Zhou Ren, and Alan Yuille.
\newblock Mitigating adversarial effects through randomization.
\newblock In {\em International Conference on Learning Representations}, 2018.
\end{thebibliography}
\end{document}